%% file: _main.tex
\input{_constants}
\arxiv % \review OR \arxiv OR \cameraready

\pdfoutput=1
\documentclass[10pt,twocolumn,letterpaper]{article}
\input{cvpr_header}

\unless\ifarxiv \myexternaldocument{_supplementary} \fi

\begin{document}
%% TITLE
\title{\paperTitle}
\author{\authorBlock}
\maketitle

\input{sec/00_abstract}

\input{sec/01_intro}

\input{sec/02_related}
\input{sec/03_background}
\input{sec/04_method}
\input{sec/05_exps}

\input{sec/06_conclusion}

{\small
\bibliographystyle{ieeenat_fullname}
\bibliography{11_references}
}

% - NEW: added title to appendix
\ifarxiv \maketitlesupplementary \fi
\ifarxiv \appendix \input{sec/07_appendix} \fi

% OLD: appendix without a title
%\ifarxiv \clearpage \appendix \input{sec/07_appendix} \fi

\end{document}

%% file: _constants.tex
\def\paperID{0000}
\def\confName{ICCV}
\def\confYear{2025}

\def\paperTitle{Benchmarking Feature Upsampling Methods for Vision Foundation Models using Interactive Segmentation}

\def\authorBlock{
    Volodymyr Havrylov\textsuperscript{1} \qquad
    Haiwen Huang\textsuperscript{1,3} \qquad
    Dan Zhang\textsuperscript{2} \qquad
    Andreas Geiger\textsuperscript{1,3} \\[0.2cm]
    \textsuperscript{1}University of Tübingen \qquad
    \textsuperscript{2}Bosch Center for Artificial Intelligence \qquad
    \textsuperscript{3}Tübingen AI Center \\[0.2cm]
}

\author{\authorBlock}

\newif\ifreview 
\newif\ifarxiv \newcommand{\arxiv}{\arxivtrue}
\newif\ifcamera 
\newif\ifrebuttal 

%% file: cvpr_header.tex
\ifreview \usepackage[review]{cvpr} \fi
\ifarxiv \usepackage[pagenumbers]{cvpr} \fi
\ifrebuttal \usepackage[rebuttal]{cvpr} \fi
\ifcamera \usepackage{cvpr} \fi

\input{_macros}  % Add packages to _macros.tex

\usepackage{xr-hyper}

\makeatletter
\newcommand*{\addFileDependency}[1]{
  \typeout{(#1)}
  \@addtofilelist{#1}
  \IfFileExists{#1}{}{\typeout{No file #1.}}
}

\makeatother
\newcommand*{\myexternaldocument}[1]{
    \externaldocument{#1}
    \addFileDependency{#1.tex}
    \addFileDependency{#1.aux}
}

\definecolor{cvprblue}{rgb}{0.21,0.49,0.74}
\usepackage[pagebackref,breaklinks,colorlinks,allcolors=cvprblue]{hyperref}
\usepackage[capitalize]{cleveref}
\crefname{section}{Sec.}{Secs.}
\crefname{table}{Table}{Tables}
\crefname{figure}{Fig.}{Figs.}

\ifarxiv \crefname{appendix}{App.}{Apps.}
\else \crefname{appendix}{Suppl.}{Suppls.} \fi

%% file: _macros.tex
\usepackage{graphicx}	
\usepackage{amsmath}	
\usepackage{amssymb}	
\usepackage{booktabs}
\usepackage{times}
\usepackage{microtype}
\usepackage{epsfig}
\usepackage{caption}
\usepackage{float}
\usepackage{placeins}
\usepackage{color, colortbl}
\usepackage{stfloats}
\usepackage{enumitem}
\usepackage{tabularx}
\usepackage{xstring}
\usepackage{multirow}
\usepackage{xspace}
\usepackage{url}
\usepackage{subcaption}
\usepackage{xcolor}
\usepackage[hang,flushmargin]{footmisc}

\ifcamera \usepackage[accsupp]{axessibility} \fi

% ...

\ifarxiv  \fi

\newcommand{\R}[1]{{%
    \textbf{%
        \ifstrequal{#1}{1}{\textcolor{red}{R#1}}{%
        \ifstrequal{#1}{2}{\textcolor{blue}{R#1}}{%
        \ifstrequal{#1}{3}{\textcolor{magenta}{R#1}}{%
        \ifstrequal{#1}{4}{\textcolor{teal}{R#1}}{%
                           \textcolor{cyan}{R#1}%
        }}}}%
    }%
}}

\usepackage{highlight}
\usepackage{colortbl}
\usepackage{multirow}
\usepackage{graphicx}

%% file: sec/00_abstract.tex
\begin{abstract}
% Abstract goes here.
Vision Foundation Models (VFMs) are large-scale, pre-trained models that serve as general-purpose backbones for various computer vision tasks. As VFMs' popularity grows, there is an increasing interest in understanding their effectiveness for dense prediction tasks. However, VFMs typically produce low-resolution features, limiting their direct applicability in this context. One way to tackle this limitation is by employing a task-agnostic feature upsampling module that refines VFM features resolution. To assess the effectiveness of this approach, we investigate Interactive Segmentation (IS) as a novel benchmark for evaluating feature upsampling methods on VFMs. Due to its inherent multimodal input, consisting of an image and a set of user-defined clicks, as well as its dense mask output, IS creates a challenging environment that demands comprehensive visual scene understanding. Our benchmarking experiments show that selecting appropriate upsampling strategies significantly improves VFM features quality. The code is released at \texttt{\href{https://github.com/havrylovv/iSegProbe}{https://github.com/havrylovv/iSegProbe}}.
\input{sec/figs/architecture}

\end{abstract}

%% file: sec/figs/architecture.tex
\begin{figure*}[htp]
    \centering
    \includegraphics[height=4.25cm]{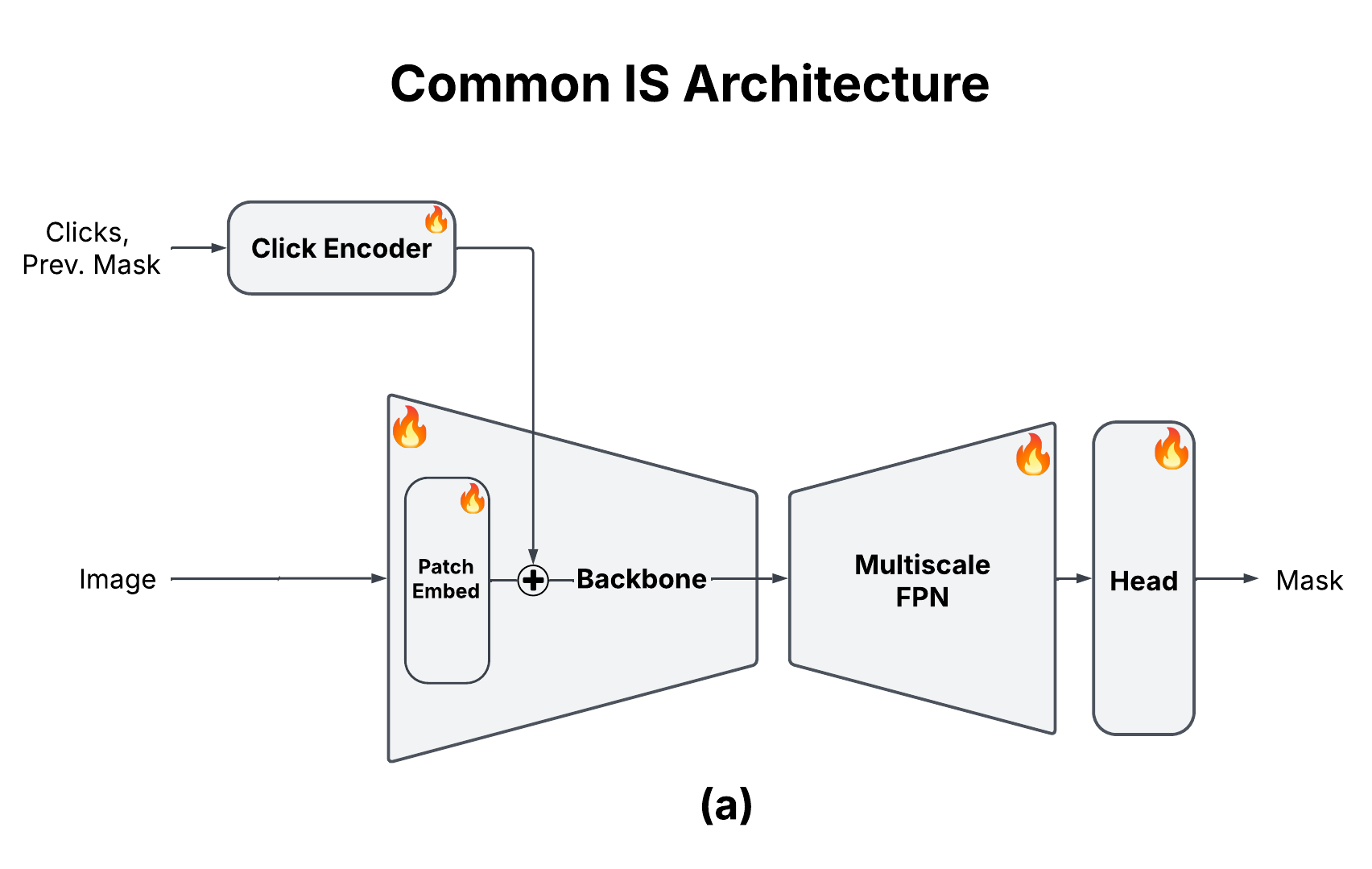}
    \hspace{0.01\textwidth} % Small space between images
    \includegraphics[height=4.25cm]{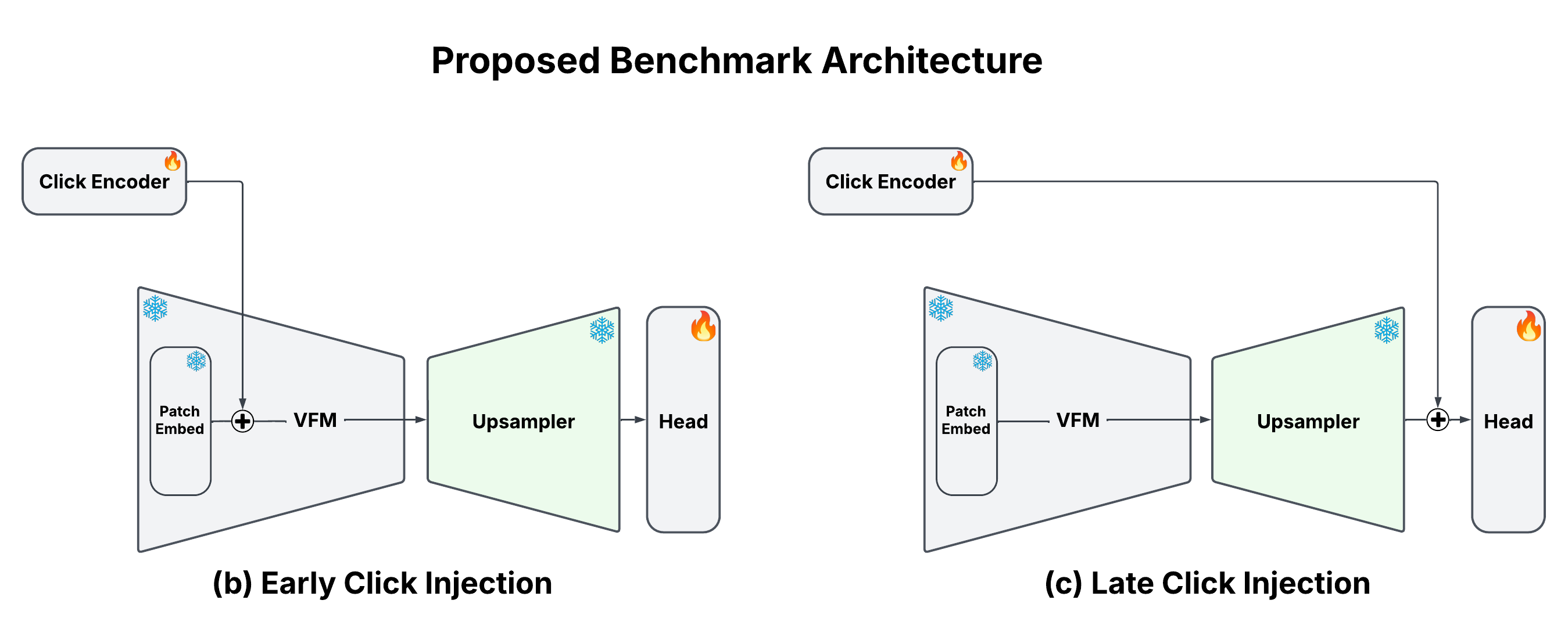}
    \caption{\textbf{General architecture of a typical IS setup (a) and the proposed benchmark (b, c).} The benchmark considers two options for injecting click features.}
    \label{fig:architecture}
\end{figure*}

%% file: sec/01_intro.tex
\section{Introduction}
\label{sec:intro}

High-quality representations from pre-trained Vision Foundation Models (VFMs) have become a crucial component of modern computer vision architectures, ensuring robustness and generalizability across various tasks \cite{kirillov2023segany, DBLP:conf/icml/RadfordKHRGASAM21, DBLP:journals/corr/abs-2502-14786, DBLP:conf/nips/YuHDSC23, DBLP:journals/corr/abs-2408-00714, DBLP:conf/nips/HuangPZ024, yuan2024ovsam, Waldmann2025PanopticonAA, Cui2024SurgicalDINOAL}. However, due to the patchification or extensive pooling operations, VFM features typically have a spatial resolution that is 16 or more times smaller than the input image.  This limits their effectiveness for dense prediction tasks, where high-resolution reasoning and fine-grained detail preservation are essential. The issue of insufficient feature resolution is commonly addressed by training a task-specific decoder, a multi-layer architecture designed for upsampling \cite{kirillov2023segany, DBLP:conf/eccv/LiMGH22, DBLP:conf/iccv/0008XBN23, DBLP:conf/nips/YuHDSC23}. Nevertheless, this strategy has several drawbacks, including high computational costs, the need to retrain the decoder for each new task, and potential data scarcity for training a high-quality decoder. A recent and promising alternative is to use a feature upsampler module \cite{DBLP:conf/iccv/WangCX0LL19, DBLP:conf/cvpr/KirillovWHG20, DBLP:conf/nips/LuLYFLC22, DBLP:conf/eccv/0003LF022, DBLP:conf/eccv/HuCXBCPW22, DBLP:conf/eccv/HuCXBCPW22, DBLP:conf/eccv/SuriWGS24, DBLP:conf/iclr/FuHBFZF24} that optionally conditions on the input image and reconstructs high-resolution features, restoring lost spatial information. As a new trend, these modules are being increasingly designed to generalize across various model configurations and downstream applications \cite{DBLP:conf/eccv/0003LF022, DBLP:conf/eccv/SuriWGS24, DBLP:conf/iclr/FuHBFZF24}. Integrating feature upsampler modules into existing pipelines has been shown to improve performance and explainability \cite{DBLP:conf/iclr/FuHBFZF24}. Moreover, combining them with a lightweight, task-specific decoder provides an alternative to using a conventional, heavy decoder when computational resources are limited.

To properly evaluate feature upsamplers on VFMs, we explore Interactive Segmentation (IS) as a novel benchmark task. IS generates object masks based on user input, where positive and negative clicks define which regions should be included or excluded from the final prediction. We specifically focus on IS due to three key characteristics: its demand for fine-grained semantic understanding, its integration of an additional input modality, and a wide range of practical applications. As a representative of 2D dense prediction tasks \cite{DBLP:conf/nips/XieWYAAL21, DBLP:conf/cvpr/GodardAB17, DBLP:conf/cvpr/WangFG15, DBLP:conf/nips/YuHDSC23}, IS poses inherent challenges for pixel-level scene understanding. Furthermore, an additional input modality in the form of sparse geometric clues (user clicks) imposes specific requirements on the architecture and allows VFMs to be tested in a multimodal regime. This setup is reminiscent of the open-vocabulary segmentation \cite{DBLP:conf/nips/YuHDSC23, DBLP:conf/nips/HuangPZ024, yuan2024ovsam}, where a textual input, instead of a spatial one, guides the predictions. Finally, IS plays a significant role in Human-Computer Interaction \cite{DBLP:journals/ijcv/KohliNRR12, Amrehn2019ASU, ramkumar2017using}, with applications including accelerated pixel-level annotation \cite{benenson2019large, marinov2024deep, TagLab}, controllable image generation \cite{zhang2023adding, DBLP:conf/cvpr/RombachBLEO22, DBLP:conf/eccv/LiYKWWXC24}, and image editing \cite{DBLP:conf/cvpr/BrooksHE23, DBLP:conf/iclr/CouaironVSC23, brack2024ledits++}. Moreover, with the increasing popularity of Segment Anything Model \cite{kirillov2023segany}, the interest in IS applications continues to grow. 

In this work, we systematically explore the design space for a feature upsampling evaluation pipeline on VFMs within the IS framework and propose a concrete architecture for this task. Specifically, we retain the VFM as a backbone, followed by an upsampler and a lightweight segmentation head. Additionally, we define modules for encoding user clicks (hereafter referred to as click encoders) and determine an appropriate strategy for integrating their features with the VFM output. Depending on the complexity of the click encoder, its features are injected into the pipeline either before or after the VFM.  
Compared to the typical IS model architecture shown in Fig. \ref{fig:architecture} (a), our approach eliminates multiscale feature extraction via Feature Pyramid Networks (FPN) and maintains a simplified segmentation head. During the experiments, we freeze the VFM and the feature upsampler, optimizing only the click encoder and segmentation head. This configuration significantly reduces both training time and computational requirements. Moreover, it more accurately reflects the effectiveness of feature upsampling. Using this architecture, we benchmark various feature upsampling techniques to identify the most effective methods for enhancing VFM features in this context. The experiments on four standard IS datasets demonstrate that the LoftUp \cite{huang2025loftuplearningcoordinatebasedfeature} upsampler yields up to a 50\% performance improvement over conventional bilinear interpolation. Furthermore, as shown in Fig. \ref{fig:features_comparison}, the upsampled features present improved sharpness, preserving fine-grained details from the input image. Finally, by establishing IS as a benchmark, we emphasize its potential to serve as a structured and effective framework for evaluating models in real-world interactive scenarios.
%%%%%%%%%%

%% file: sec/02_related.tex
\section{Related Work}
\label{sec:related}

\textbf{Interactive Segmentation.} IS refers to the task of extracting object masks based on a limited number of user clicks, which serve as a guidance. Before the advent of deep learning, IS is primarily addressed using low-level image features (e.g., intensity) in combination with graph-based optimization techniques \cite{DBLP:conf/iccv/BoykovJ01, DBLP:conf/emmcvpr/BoykovK01, DBLP:journals/pami/Grady06, DBLP:conf/cvpr/GulshanRCBZ10,
DBLP:journals/tog/RotherKB04}. With the rise of deep learning, DIOS \cite{DBLP:conf/cvpr/XuPCYH16} becomes the first method to apply neural networks to IS. DIOS also introduces a sampling strategy for generating subsequent clicks based on the current mask and formalizes training and evaluation protocols that have since become the de facto standard for click-based IS.
FCA-Net \cite{DBLP:conf/cvpr/LinZCCL20} highlights the higher impact of the first click compared to subsequent ones. RITM \cite{DBLP:conf/icip/Sofiiuk0022} removes computationally expensive inference-time optimization schemes and realizes an iterative sampling strategy during training instead. 

More recent approaches incorporate Transformer-based architectures to enhance IS performance.
FocalClick \cite{DBLP:conf/cvpr/ChenZZDQZ22} and iSegFormer \cite{DBLP:journals/corr/abs-2112-11325} leverage ViTs \cite{DBLP:conf/nips/XieWYAAL21, DBLP:conf/iccv/LiuL00W0LG21} as backbones. To further enhance efficiency and mask quality, FocalClick \cite{DBLP:conf/cvpr/ChenZZDQZ22} and FocusCut \cite{DBLP:conf/cvpr/0005DZGC22} concentrate on refining masks in local regions and its further fusion with global contextual information. SimpleClick \cite{DBLP:conf/iccv/0008XBN23} is the first model to utilize a plain, non-hierarchical ViT as a backbone. It also introduces a symmetric patch embedding layer that encodes clicks without disrupting the backbone’s pre-trained capabilities, resulting in a significant performance boost through its simplified design.

Previously described methods adopt a single-granularity approach, assuming that a click corresponds to a fixed object scale and disregarding spatial ambiguity related to object size. In contrast, the multi-granularity method GraCo \cite{DBLP:conf/cvpr/Zhao0CQZJL0C24} enables fine-grained control over output granularity by introducing an additional parameter into the input, allowing for more precise segmentation across varying object sizes.

\medskip 
\noindent \textbf{Feature Upsampling.} 
Feature upsampling is the process of increasing the spatial resolution of feature maps, typically to match the resolution of the input image. Traditional non-learnable methods, such as bilinear, bicubic, and nearest-neighbor interpolation, are commonly used for this task. However, these approaches become ineffective for large upsampling factors, as they fail to preserve fine-grained structures and often introduce blurring artifacts. To mitigate this issue, Joint Bilateral Upsampling (JBU) \cite{DBLP:journals/tog/KopfCLU07} leverages a high-resolution guidance signal, such as the input image, to refine the upsampling process.

Basic learnable upsampling methods include deconvolution \cite{DBLP:journals/corr/DumoulinV16, DBLP:conf/iccv/NohHH15, DBLP:journals/corr/ShiCTHALW16} and resize-convolution \cite{odena2016deconvolution}, both of which apply a single upsampling operator to the entire feature map. With the advancement of deep learning, a wide range of task-specific learnable upsamplers have been proposed, tailored to different model architectures and downstream applications. For instance, PointRend \cite{DBLP:conf/cvpr/KirillovWHG20} introduces a point-based rendering approach specifically for upsampling segmentation outputs. Methods such as Index Networks \cite{DBLP:journals/pami/LuDSX22} and A2U \cite{DBLP:conf/cvpr/Dai0S21} are effective primarily for image matting tasks. Other approaches, including CARAFE \cite{DBLP:conf/iccv/WangCX0LL19}, SAPA \cite{DBLP:conf/nips/LuLYFLC22}, and FADE \cite{DBLP:conf/eccv/0003LF022}, focus on predicting adaptive upsampling kernels for models following a clear encoder-decoder structure. Additionally, IFA \cite{DBLP:conf/eccv/HuCXBCPW22}, designed for semantic segmentation, constructs an implicit neural representation to align multi-level feature maps.

With the growing momentum of VFMs such as DINOv2 \cite{DBLP:journals/tmlr/OquabDMVSKFHMEA24} and CLIP \cite{DBLP:conf/icml/RadfordKHRGASAM21}, feature upsamplers are increasingly designed to be agnostic to specific downstream applications. This trend reflects the inherent versatility of VFMs, enabling upsamplers to be applied to their features across various downstream tasks. FeatUp \cite{DBLP:conf/iclr/FuHBFZF24} and LiFT \cite{DBLP:conf/eccv/SuriWGS24} are the first to suggest a task-agnostic training pipeline for feature upsamplers, demonstrating its positive impact on performance across various tasks. More recently, inspired by coordinate-based methods in 3D reconstruction, LoftUp \cite{huang2025loftuplearningcoordinatebasedfeature} introduces a coordinate-based cross-attention transformer, trained on high-resolution pseudo-GT feature maps. 

\input{sec/figs/features_comparison}

%% file: sec/figs/features_comparison.tex
% Use figure* for multi-column figure
\begin{figure*}[tp]
    \centering
    \includegraphics[width=0.9\linewidth]{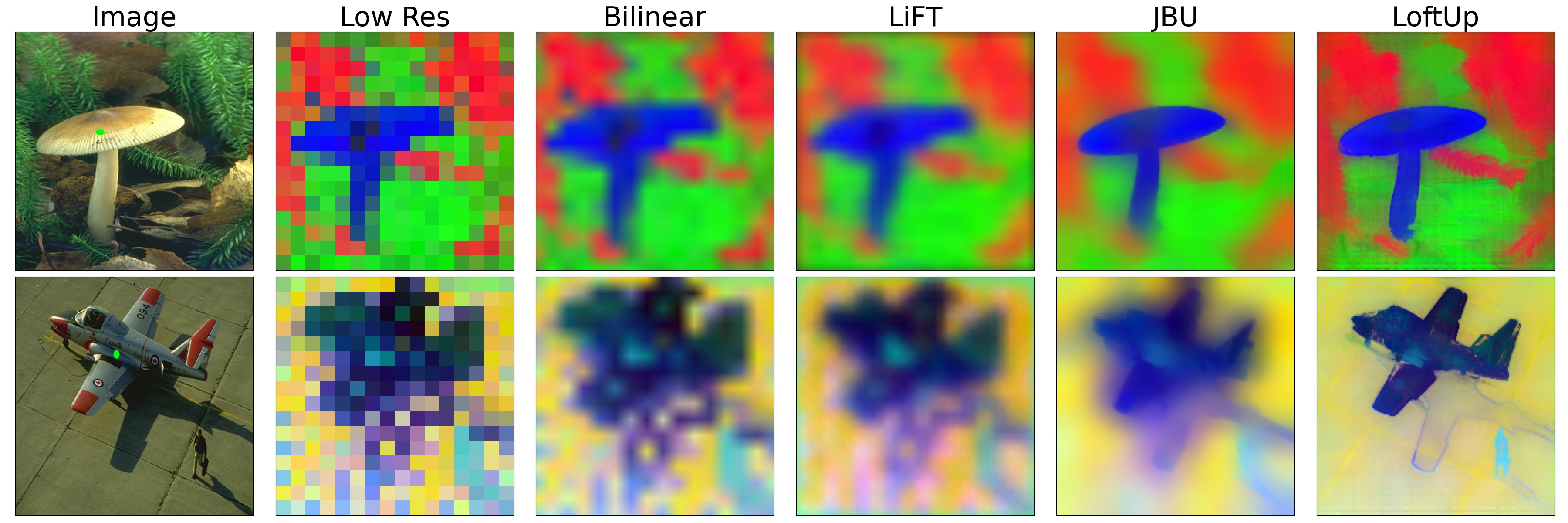}
\caption{\textbf{Comparison of features from upsamplers with a single input click.} The click is indicated by a green dot on the original images. Backbone is DINOv2 (S/14) \cite{DBLP:journals/tmlr/OquabDMVSKFHMEA24}. Click encoder is a symmetric patch embedding with early injection. Visualization method follows the PCA-based approach introduced in FeatUp \cite{DBLP:conf/iclr/FuHBFZF24}.}
    \label{fig:features_comparison}
\end{figure*}

%% file: sec/03_background.tex
\section{Interactive Segmentation Background}
\label{sec:background}

\input{sec/tables/results_main}

\input{sec/figs/is_inference_example}

\textbf{General Architecture.} State-of-the-art (SOTA) IS models \cite{DBLP:conf/iccv/0008XBN23, DBLP:conf/cvpr/Zhao0CQZJL0C24} typically consist of four key components, as shown in Fig. \ref{fig:architecture} (a):  
\begin{enumerate}
    \item \textbf{Click encoder:} converts sparse user interactions (clicks) into dense features, referred to as click features. 
    \item \textbf{Backbone:} encodes the input image along with the click features into unified feature representations.  
    \item \textbf{Multiscale feature pyramid network (FPN):} refines feature representations across different scales.  
    \item \textbf{Multiscale segmentation head:} aggregates features from different levels of the FPN to generate the final segmentation mask.
\end{enumerate}

In this architecture, the FPN's role is essentially to learn a task-specific feature upsampler for better dense prediction. For instance, SimpleClick \cite{DBLP:conf/iccv/0008XBN23} incorporates an FPN adapted from ViTDet \cite{DBLP:conf/eccv/LiMGH22}, which is designed for plain vision backbones. This FPN generates multiscale feature maps based solely on the output of the backbone’s final layer. To process the resulting features, SimpleClick further employs an adapted SegFormer head \cite{DBLP:conf/nips/XieWYAAL21}, which aligns features to a common resolution and fuses them for final mask prediction.

The click encoder is a distinctive element of IS and necessitates special attention. The primary function of the click encoder is to process a user-defined set of positive and negative clicks, transforming them into a feature representation. These features are then integrated into the IS model to provide precise guidance during mask generation. A typical click encoding pipeline \cite{DBLP:conf/iccv/0008XBN23} is described as follows. As a preprocessing step, user clicks are first represented as a two-channel disk map, where the first channel is assigned to positive clicks and the second to negative ones. To enhance guidance, the segmentation mask from the previous step is incorporated as a third channel. This map maintains the same resolution as the input image and serves as the direct input to the click encoder. Fig. \ref{fig:is_inference_example} illustrates a concrete example of such an input representation. \textit{Symmetric patch embedding} module \cite{DBLP:conf/iccv/0008XBN23} acts as a click encoder and is designed to replicate the patch embedding layer typically used at the beginning of vision backbones. It consists of a single convolutional layer that partitions the input into patches and linearly projects them into a sequence of feature vectors. Resulting features are added element-wise to intermediate image features immediately after the patch embedding layer of the vision backbone. This type of feature injection will be further referred to as \textit{early injection}.

\medskip
\noindent \textbf{Training and Evaluation Protocols.} 
 The principles for training and evaluation in IS are well-established and widely adopted across most studies in the field \cite{DBLP:conf/iccv/0008XBN23, DBLP:conf/cvpr/Zhao0CQZJL0C24, DBLP:conf/icip/Sofiiuk0022, DBLP:conf/cvpr/ChenZZDQZ22, DBLP:conf/cvpr/LinZCCL20, benenson2019large, DBLP:conf/cvpr/XuPCYH16}. The key aspect that distinguishes IS from other segmentation tasks is its click generation process.

During evaluation, we use the approach from \cite{DBLP:conf/cvpr/LiCK18, DBLP:conf/cvpr/XuPCYH16} to place user clicks in a way that mimics natural human behavior. Specifically, each successive click is placed at the center of the largest region that contains any type of prediction error (including both false positive and false negative). The center point is the furthest point from the region's boundaries. This method of click creation, however, is unsuitable for the training phase due to its deterministic nature, which limits click diversity.

To address this, we generate user clicks during training by following the protocol proposed in RITM \cite{DBLP:conf/icip/Sofiiuk0022}. Here, clicks are automatically simulated using two strategies: random and iterative. First, the random generation strategy places the clicks on the target object at random, thereby increasing the variety of click locations. Then, similar to the evaluation procedure, the iterative strategy sequentially places clicks in regions where previous predictions exhibit errors. To avoid overfitting, the iterative strategy does not directly select the center of a mislabelled region as the next click. Instead, it first applies a morphological erosion operation to reduce the region's area by 4 times. 

%% file: sec/tables/results_main.tex
\begin{table*}[h]
    \centering
    \renewcommand{\arraystretch}{1.2} % Reduce row height
    \setlength{\tabcolsep}{3.5pt} % Reduce column separation
    \resizebox{\textwidth}{!}{
    \begin{tabular}{>{\raggedright\arraybackslash}m{1.7 cm}|cccc!{\color{gray!50}\vrule}cccc!{\color{gray!50}\vrule}cccc!{\color{gray!50}\vrule}cccc}
        \toprule
        \multirow{2}{*}{\centering Method} & \multicolumn{4}{c}{GrabCut} & \multicolumn{4}{c}{Berkeley} & \multicolumn{4}{c}{DAVIS} & \multicolumn{4}{c}{SBD} \\
        & NoC80 & NoC85 & NoC90 & IoU@1 
        & NoC80 & NoC85 & NoC90 & IoU@1 
        & NoC80 & NoC85 & NoC90 & IoU@1 
        & NoC80 & NoC85 & NoC90 & IoU@1 \\
        \midrule
        \rowcolor{gray!20} % Light gray row
        \multicolumn{17}{l}{\textit{Click Encoder: Symmetric Patch Embedding} \cite{DBLP:conf/iccv/0008XBN23}} \\
        Low-res & 4.16 & 5.72 & 10.50 & 64.90 & 5.67 & 8.93 & 14.22 & 55.77 & 10.55 & 14.23 & 17.27 & 52.82 & 7.92 & 10.60 & 14.09 & 45.80 \\
        Bilinear & 4.32 & 6.18 & 10.58 & 65.04 & 6.58 & 10.08 & 15.17 & 55.83 & 11.61 & 15.01 & 17.73 & 54.26 & 8.29 & 11.16 & 14.84 & 44.58 \\
        LiFT \cite{DBLP:conf/eccv/SuriWGS24} & 13.32 & 16.24 & 19.04 & 29.66 & 14.53 & 16.75 & 19.05 & 31.99 & 16.13 & 18.00 & 19.53 & 41.14 &  12.67 & 15.57 & 18.35 & 33.76 \\
        FeatUp \cite{DBLP:conf/iclr/FuHBFZF24} & 4.04 & 5.32 & 8.24 & 65.89 & 5.43 & 8.21 & 12.22 & 56.67 & 9.89 & 13.36 & 16.63 & 55.03 & 7.84 & 10.26 & 13.71 & 43.78 \\
        LoftUp \cite{huang2025loftuplearningcoordinatebasedfeature}& \textbf{1.72} & \textbf{2.92} & \textbf{4.66} & \textbf{78.49} & \textbf{3.04} & \textbf{4.76} & \textbf{8.69} & \textbf{65.24} & \textbf{6.37} & \textbf{9.60} & \textbf{14.25} & \textbf{67.31} & \textbf{6.30} & \textbf{8.46} & \textbf{11.87} & \textbf{50.22} \\
        \rowcolor{gray!20} % Light gray row
        \multicolumn{17}{l}{\textit{Click Encoder: SimpleViT} \cite{DBLP:journals/corr/abs-2205-01580} } \\
        Low-res & 3.10 & 4.04 & 7.54 & 65.55 & 4.26 & 6.28 & 12.53 & 65.65 & 7.52 & 11.36 & 15.83 & 61.07 & 6.55 & 8.91 & 12.44 & 52.37  \\
        Bilinear & 3.38 & 4.92 & 8.68 & 64.40 & 4.63 & 7.45 & 13.72 & 64.11 & 7.94 & 12.02 & 16.50 & 60.63 & 6.80 & 9.53 & 13.36 & 49.67 \\
        LiFT \cite{DBLP:conf/eccv/SuriWGS24} & 3.64 & 4.78 & 7.64 & 64.17 & 4.12 & 6.33 & 11.87 & 62.82 & 7.31 & 10.70 & 15.83 & 58.86 & 6.57 & 8.94 & 12.53 & 48.68 \\
        FeatUp \cite{DBLP:conf/iclr/FuHBFZF24} & 2.28 & 2.76 & 5.78 & 71.46 & 3.46 & 5.92 & 11.00 & 66.70 & 6.43 & 10.08 & 15.29 & 61.87 & 7.29 & 9.64 & 13.08 & 47.37 \\
        LoftUp \cite{huang2025loftuplearningcoordinatebasedfeature} & \textbf{1.92} & \textbf{2.54} & \textbf{3.76} & \textbf{72.68} & \textbf{2.19} & \textbf{3.07} & \textbf{5.42} & \textbf{70.55} & \textbf{3.55} & \textbf{5.33} & \textbf{10.14} & \textbf{71.27} & \textbf{4.67} & \textbf{6.39} & \textbf{9.79} & \textbf{58.42} \\
        \bottomrule
    \end{tabular}
    }
    \caption{\textbf{Comparison of upsamplers for two click encoders. Adding an appropriate upsampler module generally improves model performance.} The vision backbone is DINOv2 (S/14) \cite{DBLP:journals/tmlr/OquabDMVSKFHMEA24}, and the segmentation head consists of two $3 \times 3$ convolutions, followed by a single $1 \times 1$ convolution. The symmetric patch embedding encoder \cite{DBLP:conf/iccv/0008XBN23} employs early click feature injection, whereas SimpleViT \cite{DBLP:journals/corr/abs-2205-01580} uses late injection.}

    \label{tab:main_results}
\end{table*}

% The model comprises a vision backbone, a click encoder, an upsampler, and a segmentation head. Models are trained on the SBD dataset \cite{DBLP:conf/iccv/HariharanABMM11} for 20 epochs with a batch size of 8 and an image resolution of $224 \times 224$. 

%% file: sec/figs/is_inference_example.tex
\begin{figure}[htp]
    \centering
    \includegraphics[width=0.45\textwidth]{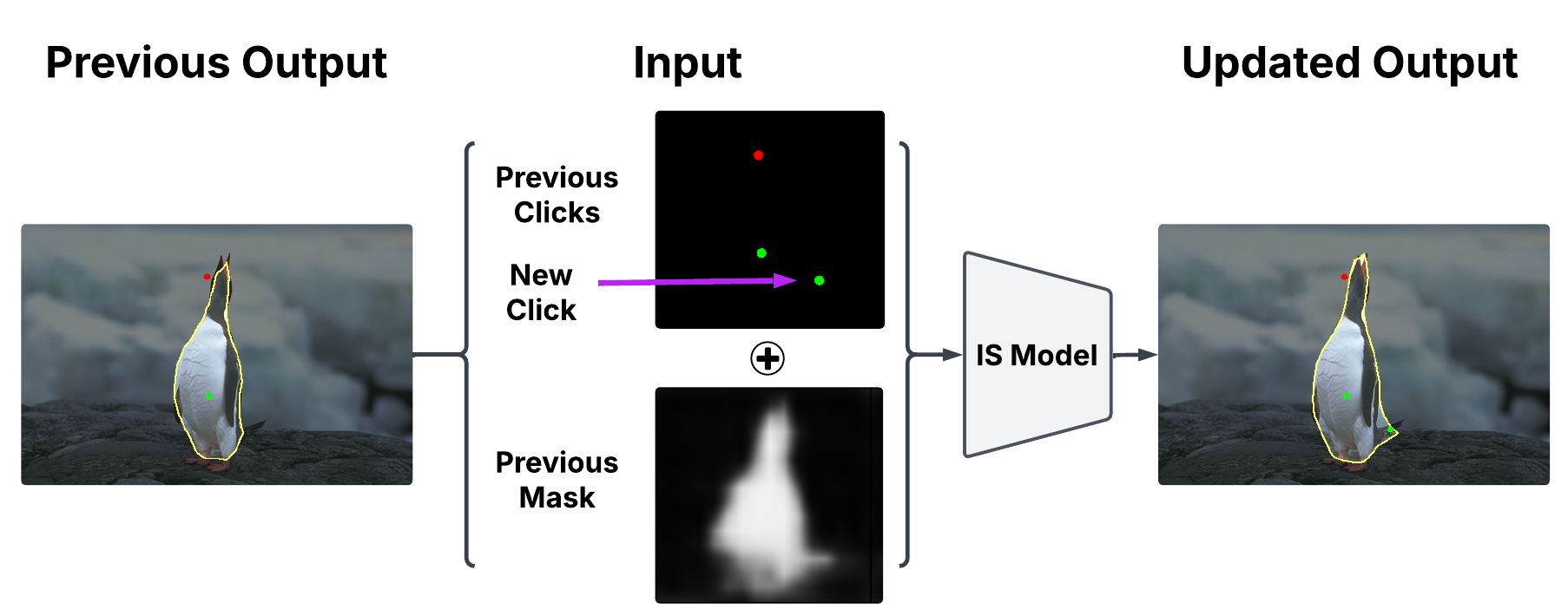}
    \caption{\textbf{Detailed example of IS inference.} The model takes an input image along with a stacked representation of two disk maps indicating positive and negative clicks. Positive and negative clicks are indicated by green and red dots, respectively. Additionally, the model may receive a probability map from the previous iteration. The input image is not shown for clarity.}
    \label{fig:is_inference_example}
\end{figure}

%% file: sec/04_method.tex
\section{Proposed Benchmark}
\label{sec:method}

This study establishes a benchmarking framework for feature upsamplers on VFMs within the IS task. 
Drawing inspiration from linear probing \cite{DBLP:journals/corr/AlainB16, Thrush2022WinogroundPV, DBLP:journals/corr/abs-2412-09606, Aydemir2024CanVF, ElBanani2024ProbingT3, Ranzinger2023AMRADIOAV}, we freeze both the backbone and the feature upsampler throughout all experiments. This design choice not only isolates and highlights the contribution of the upsampler but also significantly reduces training time and computational requirements. 
As the primary role of the feature upsampler is to restore the spatial resolution lost due to downsampling, its inclusion eliminates the need to learn feature upsampling through an FPN -- a strategy commonly adopted by current SOTA methods \cite{DBLP:conf/iccv/0008XBN23, DBLP:conf/cvpr/Zhao0CQZJL0C24}. Therefore, in our setup, the VFM is followed by the upsampler and a lightweight, single-scale segmentation head, as illustrated in Fig. \ref{fig:architecture} (b, c). While traditional linear probing employs a single-layer linear head, this approach proved insufficient due to the inherent complexity of IS tasks. To ensure a more meaningful comparison with SOTA methods, we extend the segmentation head to three layers. Further details on the segmentation head design and selection can be found in Sec. \ref{appendix:seg_head_ablation}.

Beyond the widely used \textit{symmetric patch embedding} click encoder with \textit{early injection} \cite{DBLP:conf/iccv/0008XBN23, DBLP:conf/cvpr/ChenZZDQZ22}, we explore an alternative click encoder and injection mechanism. The second, more powerful click encoder is \textit{SimpleViT} \cite{DBLP:journals/corr/abs-2205-01580}, an improved variant of the vanilla ViT \cite{DBLP:conf/iclr/DosovitskiyB0WZ21}. It was selected primarily for its faster convergence and superior performance compared to its predecessor. Additionally, we introduce \textit{late injection}, where click features are element-wise added to the final image features after the upsampler. In both \textit{early} and \textit{late} injection schemes, image and click features shapes are matched to ensure proper addition.

\noindent \textbf{Other Explorations.} Several additional design choices were explored but proved unsuccessful. These included experimenting with upsamplers to construct a multiscale FPN \cite{DBLP:conf/eccv/LiMGH22} and incorporating a granularity input parameter, as suggested by GraCo \cite{DBLP:conf/cvpr/Zhao0CQZJL0C24}. Further details on these efforts are provided in Sec. \ref{sec:alternative_is_benchmarks}.

%% file: sec/05_exps.tex
\section{Experiments}
\label{sec:exps}

\subsection{Experimental Setup}
\input{sec/figs/viz_masks_positive}
\medskip
\noindent \textbf{Features.} Our primary backbone of interest is DINOv2 \cite{DBLP:journals/tmlr/OquabDMVSKFHMEA24} (ViT-S/14), but we also conduct ablation studies with ViT \cite{DBLP:conf/iclr/DosovitskiyB0WZ21}. For all experiments, publicly available checkpoints are used. 

\medskip
\noindent \textbf{Upsamplers.} We evaluate five different feature upsampling strategies. As a baseline, we consider a case without explicit upsampling, where the segmentation head is trained directly on the low-resolution backbone features. The resulting low-resolution segmentation mask is then upsampled to the input resolution using bilinear interpolation.

Among non-learnable upsamplers, we utilize bilinear interpolation, which is applied immediately after the backbone to scale feature maps to the input resolution, enabling the segmentation head to operate on high-resolution features.

The remaining three feature upsampling methods fall under the category of image-adaptive upsampling, as they leverage the input image as guidance to preserve high-frequency details. These methods include  LiFT \cite{DBLP:conf/eccv/SuriWGS24}, FeatUp’s JBU \cite{DBLP:conf/iclr/FuHBFZF24} and LoftUp \cite{huang2025loftuplearningcoordinatebasedfeature}. LiFT and FeatUp’s JBU upscale feature maps by fixed factors of 2 and 16, respectively. To ensure consistent evaluation across architectures, bilinear interpolation is applied to adjust the upsampled feature maps to match the input image resolution. In contrast, LoftUp directly upsamples the feature maps to an arbitrary resolution defined by the input image, eliminating the need for additional interpolation.

\medskip 
\noindent \textbf{Datasets.} Our experiments are conducted on 4 widely used IS datasets, detailed as follows: 
\begin{itemize}
    \item \textbf{GrabCut} \cite{DBLP:journals/tog/RotherKB04}: contains 50 images, each with a single, well-defined object instance.
    \item \textbf{Berkeley} \cite{DBLP:conf/iccv/MartinFTM01}: contains 96 images with 100 instances of greater complexity. It partially overlaps with GrabCut. 
    \item \textbf{DAVIS} \cite{DBLP:conf/cvpr/PerazziPMGGS16}: contains 50 high-quality video sequences. Following prior works \cite{DBLP:conf/cvpr/Zhao0CQZJL0C24,DBLP:conf/iccv/0008XBN23,DBLP:conf/cvpr/ChenZZDQZ22,DBLP:conf/icip/Sofiiuk0022}, we use a subset of 345 frames.
    \item \textbf{SBD} \cite{DBLP:conf/iccv/HariharanABMM11}: contains 8498 training images with 20172 instances and 2857 validation images with 6671 instances.
\end{itemize}

\medskip
\noindent \textbf{Training Details.} Our training pipeline follows the methodology introduced in SimpleClick \cite{DBLP:conf/iccv/0008XBN23}. Models are trained on SBD dataset for 20 epochs with the normalized focal loss \cite{DBLP:conf/icip/Sofiiuk0022}, which has been shown to accelerate convergence and improve performance compared to binary cross-entropy loss. Training image resolution is set to $224 \times 224$, and the same data augmentation techniques as in \cite{DBLP:conf/iccv/0008XBN23} are applied.
Optimization is performed using Adam with $\beta_1 = 0.9$ and $\beta_2=0.999$. The initial learning rate is set to $5 \times 10^{-5}$ and is reduced to $5 \times 10^{-6}$ after the 17th epoch. All experiments are conducted on a single NVIDIA RTX 3090 GPU with a batch size of 8 unless stated otherwise.

\medskip
\noindent \textbf{Evaluation Details.} 
Our evaluation protocol is based on prior works \cite{DBLP:conf/iccv/0008XBN23, DBLP:conf/icip/Sofiiuk0022}. Models are evaluated on four widely used datasets: GrabCut, Berkeley, DAVIS and SBD. The maximum number of clicks for each instance is fixed to 20. For performance assessment, we use the Number of Clicks (NoC) metric, which quantifies the number of user clicks required to achieve a predefined Intersection over Union (IoU) threshold. We report NoC at three IoU levels: 80\% (NoC80), 85\% (NoC85), and 90\% (NoC90). Additionally, we employ the IoU@$k$ metric, which evaluates the segmentation quality after a fixed number of $k$ user clicks.

\subsection{Findings}
\input{sec/figs/viz_masks_negative}

\textbf{Optimal Click Injection.}
An early stage of this research focused on identifying an architecture that ensures a fair and consistent comparison of different feature upsamplers within the IS task. Unlike conventional IS models, where all components are trainable, our setup keeps the backbone and upsampler frozen throughout all experiments. Given that these components, along with the segmentation head, were predefined, the primary challenge was determining an effective method for computing and injecting click features without compromising VFMs.

Experiments revealed that the widely adopted symmetric patch embedding click encoder serves as a reasonable baseline, despite having very few trainable parameters. However, it achieves optimal performance only when combined with early injection (Fig. \ref{fig:architecture}, b). In contrast, a more expressive encoder, such as SimpleViT, delivers superior results, but only with late injection (Fig. \ref{fig:architecture}, c). These optimal click feature injection strategies were utilized in all subsequent experiments. Switching injection types leads to a significant degradation in performance. One possible explanation is that using a simpler click encoder with late injection lacks sufficient representational power to effectively learn click features independently, forcing it to rely on the VFM's features. Conversely, when employing a more complex click encoder with early injection, backpropagating through the frozen VFM and upsampler may impede effective learning.

Additionally, upsampling click features separately and merging it with upsampled image features further improved performance. However, to maintain architectural simplicity, this configuration was not adopted for subsequent benchmarking. Supporting quantitative results for FeatUp’s JBU are provided in Tab. \ref{tab:optimal_click_injection}. 
\input{sec/tables/optimal_click_injection}

\input{sec/figs/ious}

\medskip
\noindent \textbf{Comparative Analysis.} 
Following the selection of the architecture, several benchmarking experiments were conducted. Tab. \ref{tab:main_results} presents a comparative analysis of the upsampling techniques discussed in Sec. \ref{sec:method} for two click encoders. Based on prior findings, we report results for the symmetric patch embedding and SimpleViT click encoders, combined with early and late injection, respectively.

The quantitative results indicate that directly training the segmentation head on low-resolution features and performing bilinear upsampling at the prediction level (Low-res case) consistently outperforms applying bilinear upsampling at the feature level, i.e., between the backbone and the segmentation head. This suggests that the early introduction of bilinear upsampling may introduce artifacts that degrade performance. Among the learnable feature upsamplers tested, LiFT performs the worst. This can be attributed to training against pseudo-GT features at low resolution, which both limits the upsampling factor to 2$\times$ and provides a weak supervision signal, potentially insufficient for learning fine-grained details. Notably, when combined with the patch embedding click encoder, LiFT shows almost no learning, indicating a possible lack of model capacity. FeatUp, which frames its training as a multi-view reconstruction proxy task, achieves better results than LiFT. However, it still struggles to learn high-resolution features from low-resolution pseudo-GT. Furthermore, LiFT and FeatUp architectures (U-Net and modified JBU, respectively) both rely solely on locally predicted kernels, limiting global interaction during upsampling. In contrast, LoftUp consistently outperforms all other methods across all metrics and datasets. The key factors contributing to this performance are the use of full-resolution pseudo-GT features during training and a cross-attention design based on a coordinate-based transformer architecture. The former provides much denser supervision for learning fine-grained feature details, while the latter enables global attention and content-aware upsampling. Overall, the reported quantitative results align with our expectations because (1) the results are consistent with the qualitative observations; and (2) LoftUp introduces significant changes to both the architecture and the training objective of feature upsamplers, addressing long-standing challenges such as GT data scarcity and limited global context. Surprisingly, the smallest gains relative to the low-resolution baseline were observed on the SBD dataset, which was also used during training. This may be attributed to the use of the largest evaluation subset, where the size difference with other datasets can reach up to two orders of magnitude. 

For qualitative evaluation, upsampled VFM's features and segmentation masks generated from a single click were visualized. Fig. \ref{fig:features_comparison} shows PCA projections of features produced by different upsampling methods. Fig. \ref{fig:viz_results_positive} presents successful cases of segmentation. These figures illustrate that the inclusion of LoftUp yields the most substantial improvements in segmentation quality.
Conversely, in cases with poor performance (Fig. \ref{fig:viz_masks_negative}), the primary causes of failure are typically the ambiguity in the GT mask scale or the incorrect placement of the initial click. In IS pipeline, the first click is automatically determined as the point farthest from the GT mask boundaries. However, for masks with irregular shapes (Fig. \ref{fig:viz_masks_negative}, first row), this click may not accurately correspond to the intended object. Additionally, some GT masks only partially cover a larger, semantically complete object, making it challenging for the model to recover fine-grained details (Fig. \ref{fig:viz_masks_negative}, second row).  However, such ambiguities can be effectively resolved by including additional user clicks.

\medskip
\noindent \textbf{Performance Saturates as Clicks Increase.} To support the claim that additional clicks can resolve ambiguities related to mask scale and click placement, Fig. \ref{fig:ious} presents the mean IoU as a function of the number of user clicks. With the exception of LiFT, which performs poorly, all upsampler configurations achieve at least $80\%$ object coverage after $20$ clicks across the four datasets studied. The convergence rate and maximum IoU attained vary depending on dataset complexity. The low-resolution baseline generally performs similarly to bilinear interpolation, while FeatUp exhibits a slight improvement. LoftUp achieves the highest performance overall.

%% file: sec/figs/viz_masks_positive.tex
% Use figure* for multi-column figure
\begin{figure*}[tp]
    \centering
    \includegraphics[width=\linewidth]{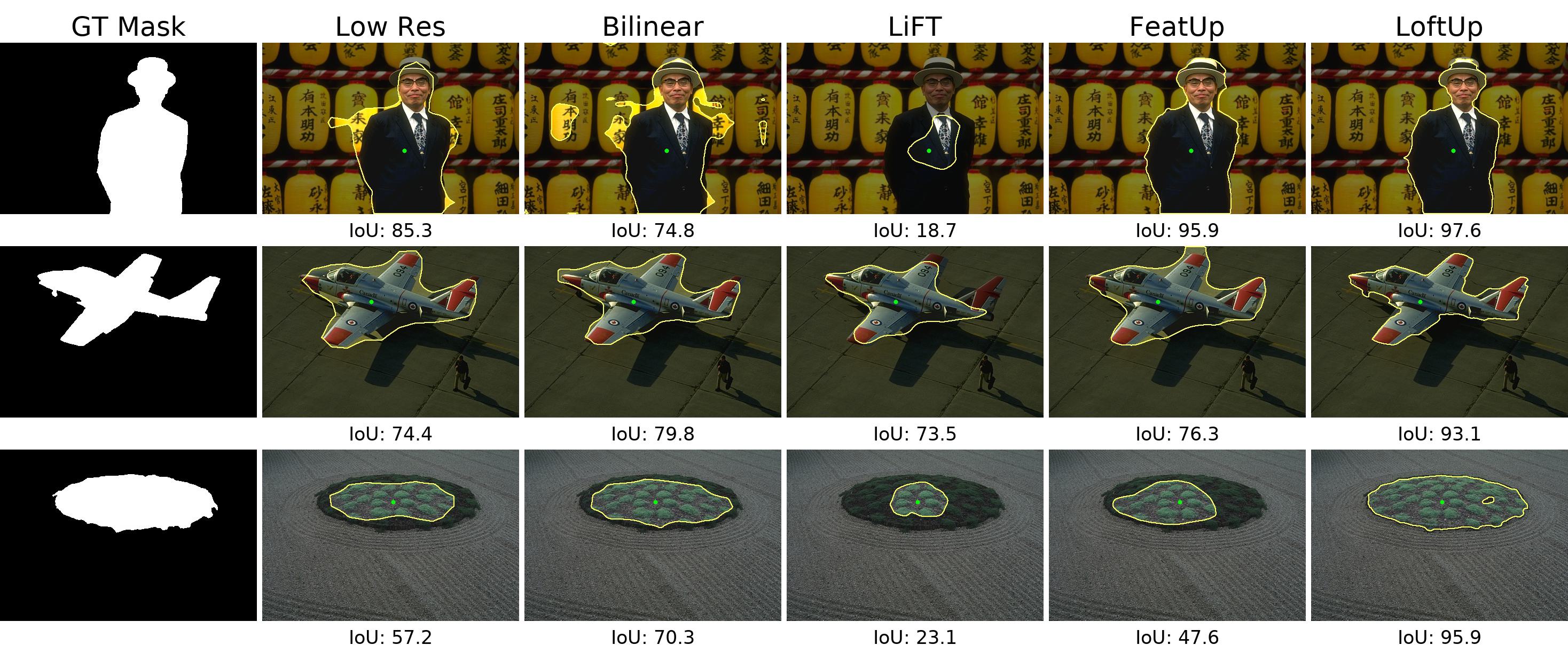}
    \caption{\textbf{Segmentation results on GrabCut} \cite{DBLP:journals/tog/RotherKB04}\textbf{ with a single input click. Successful cases.} The click is indicated by a green dot. Backbone is DINOv2 (S/14) \cite{DBLP:journals/tmlr/OquabDMVSKFHMEA24}. Click encoder is a symmetric patch embedding with early injection.}
    \label{fig:viz_results_positive}
\end{figure*}

%% file: sec/figs/viz_masks_negative.tex
% Use figure* for multi-column figure
\begin{figure*}[ht]
    \centering
    \includegraphics[width=0.8\linewidth]{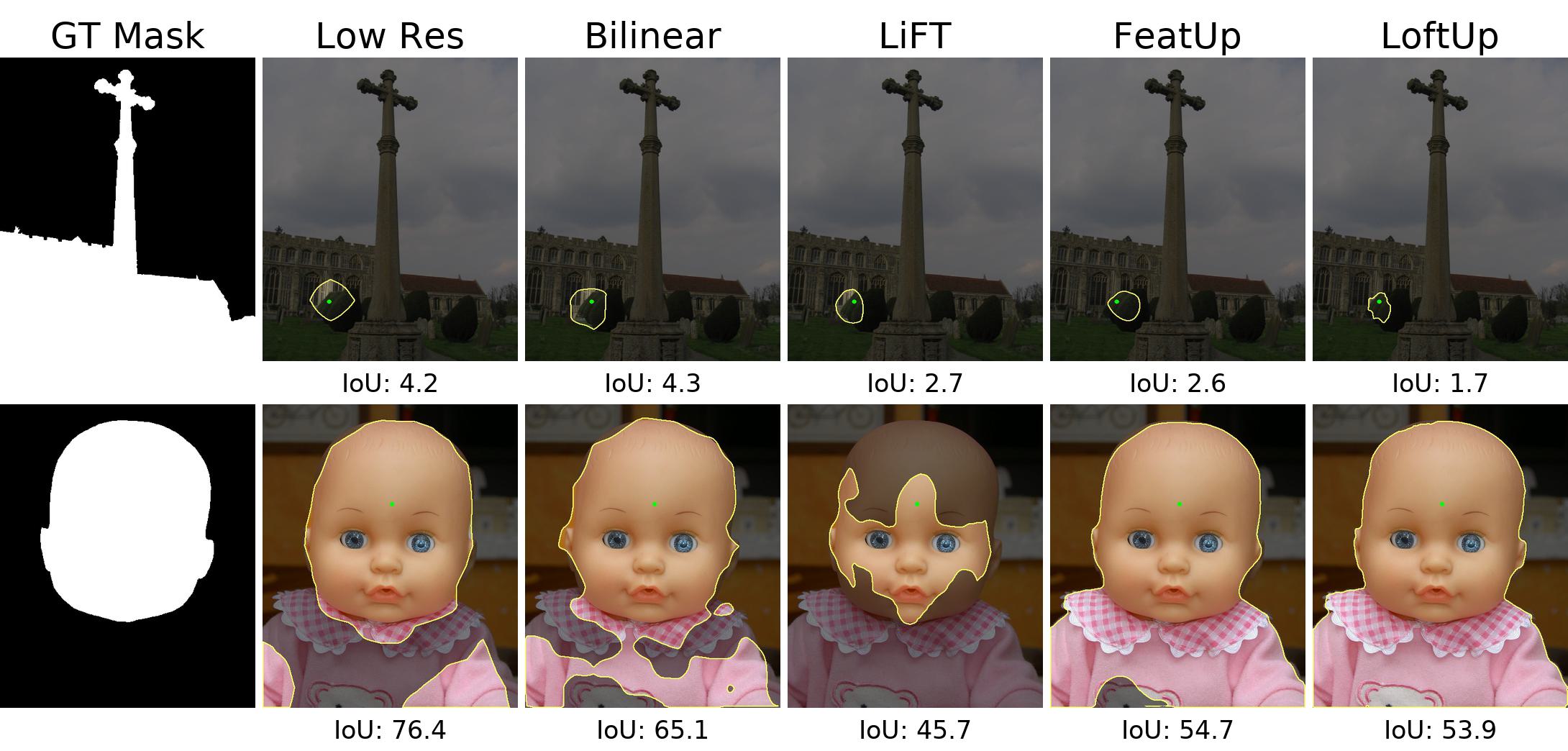}
    \caption{\textbf{Segmentation results on GrabCut} \cite{DBLP:journals/tog/RotherKB04}\textbf{ with a single input click. Failure cases.} The click is indicated by a green dot. Backbone is DINOv2 (S/14) \cite{DBLP:journals/tmlr/OquabDMVSKFHMEA24}. Click encoder is a symmetric patch embedding with early injection.}
    \label{fig:viz_masks_negative}
\end{figure*}

%% file: sec/tables/optimal_click_injection.tex
\begin{table}[h]
    \centering
    \renewcommand{\arraystretch}{1.1} % Adjust row height
    \setlength{\tabcolsep}{2.5pt} % Reduce column spacing
    \small
    \begin{tabular}{l|cccc}
        \toprule
        \multirow{2}{*}{\centering Injection Type} & \multicolumn{4}{c}{GrabCut} \\
         & NoC80 & NoC85 & NoC90 & IoU@1 \\
        \midrule
        \rowcolor{gray!20} % Light gray row
        \multicolumn{5}{l}{\textit{ Click Encoder: Symmetric Patch Embedding} \cite{DBLP:conf/iccv/0008XBN23}} \\
        Early Injection & 3.68 & 5.02 & 8.02 & 69.61 \\
        Late Injection & 6.72 & 8.28 & 11.08 & 51.67 \\
        \rowcolor{gray!20} % Light gray row
        \multicolumn{5}{l}{\textit{Click Encoder: SimpleViT} \cite{DBLP:journals/corr/abs-2205-01580}} \\
        Early Injection & 9.24 & 11.16 & 13.82 & 26.45 \\
        Late Injection & 2.72 & 3.66 & 7.34 & 69.84 \\
        After Separate Upsampling & 2.3 & 3.72 & 6.62 & 71.57 \\
        \bottomrule
    \end{tabular}

    \caption{\textbf{Comparison of click injection mechanisms.} Backbone is DINOv2 (S/14) \cite{DBLP:journals/tmlr/OquabDMVSKFHMEA24} and upsampler is FeatUp's JBU \cite{DBLP:conf/iclr/FuHBFZF24}. "After Separate Upsampling" indicates that click and image features were upsampled independently before merging.}

    \label{tab:optimal_click_injection}
\end{table}

%% file: sec/figs/ious.tex
% Use figure* for multi-column figure ht
\begin{figure*}[ht]
    \centering
    \includegraphics[width=0.6\linewidth]{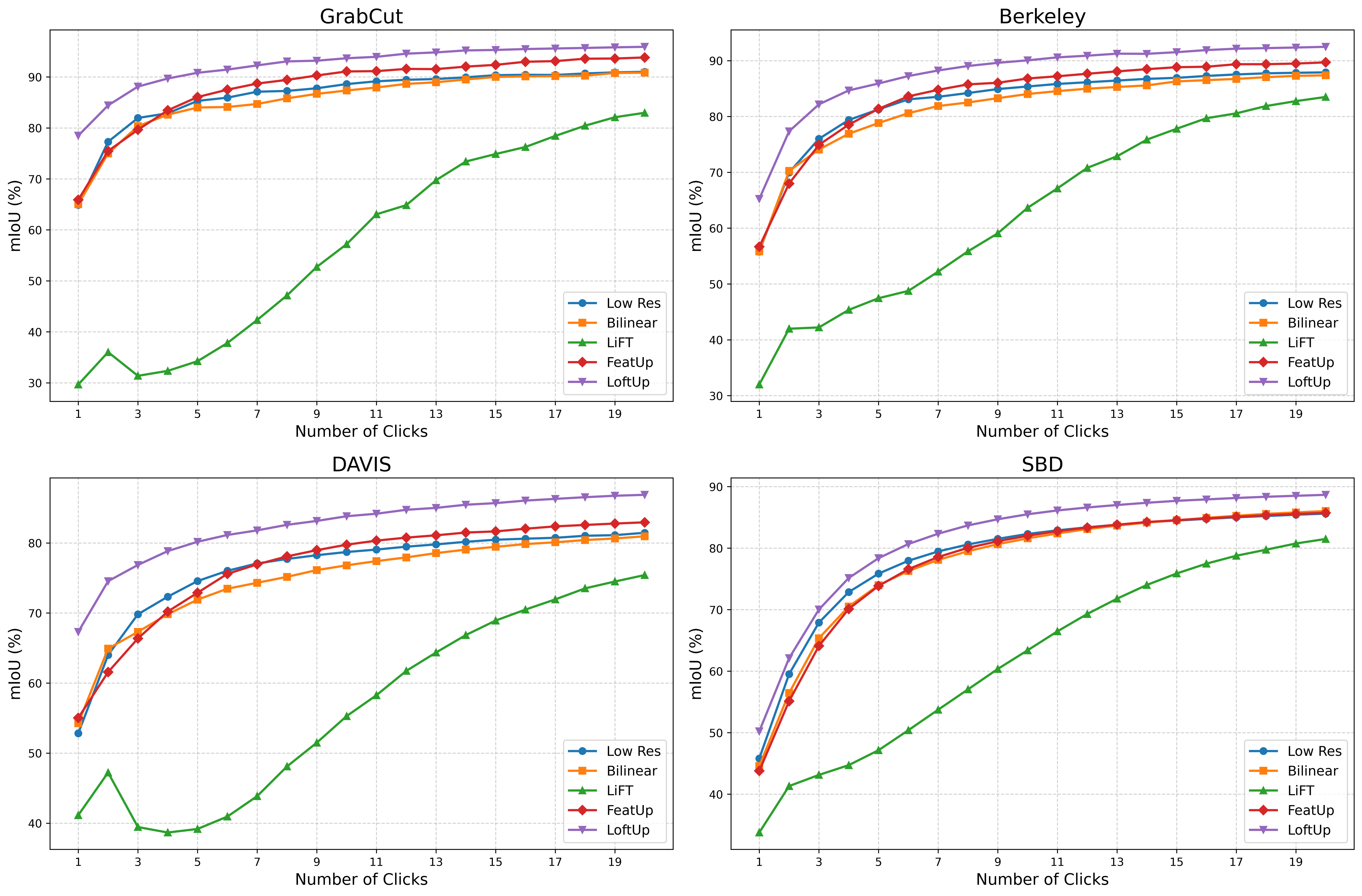}
    \caption{\textbf{Convergence of IoU with increasing user clicks.} Backbone is DINOv2 (S/14) \cite{DBLP:journals/tmlr/OquabDMVSKFHMEA24}. Click encoder is a symmetric patch embedding with early injection.}
    \label{fig:ious}
\end{figure*}

%% file: sec/06_conclusion.tex
\section{Conclusion}
\label{sec:conclusion}
In this work, we systematically investigated various configurations for evaluating feature upsamplers on VFMs within IS task and introduced a robust benchmark architecture for this purpose. Our findings demonstrate that IS is not only an effective task for evaluating the dense prediction capabilities of VFMs, but also that suitable feature upsamplers can significantly enhance overall model performance. We hope this work encourages further research on VFMs and feature upsamplers, promoting IS as a standard evaluation task.

%% file: sec/07_appendix.tex
\section{Further Architectural Explorations}

\input{sec/tables/multiscale_benchmark}
\subsection{Segmentation Head Ablation Studies}
\label{appendix:seg_head_ablation}
\noindent To determine the optimal segmentation head, three configurations were evaluated, aiming to maintain simplicity while ensuring sufficient expressiveness.

\begin{itemize}
    \item \textit{Linear Head} – A single $1 \times 1$ convolutional layer.
    \item \textit{Simple Conv Head} – Three $1 \times 1$ convolutional layers with an inner channel dimension of 384.
    \item \textit{Conv Head} – Similar to the \textit{Simple Conv Head}, but with the first two layers using a kernel size of $3 \times 3$ instead of $1 \times 1$.
\end{itemize}

Tab. \ref{tab:ablation_head} presents a comparison of these architectures using ViT \cite{DBLP:conf/iclr/DosovitskiyB0WZ21} backbone without feature upsampling, a symmetric patch embedding click encoder, and early click injection. The \textit{Conv Head} achieves the best balance between performance and complexity and is therefore selected for all subsequent experiments.

\input{sec/tables/ablation_head}

\subsection{Alternative IS Benchmarks}
\label{sec:alternative_is_benchmarks}
In this work, we explored several alternative architectures for benchmarking VFMs, none of which proved successful. Here, we briefly outline these approaches and provide our hypotheses regarding their failures.

\medskip
\noindent \textbf{Multiscale Benchmark.} SOTA IS models \cite{DBLP:conf/iccv/0008XBN23, DBLP:conf/cvpr/Zhao0CQZJL0C24} typically leverage multiscale features, which are extracted from vision backbone outputs via FPN and subsequently processed by a multiscale segmentation head. While our primary benchmark focused on removing all multiscale components, here we investigate whether upsamplers can be beneficially used in FPNs.  

The architecture of our \textit{Upsampler-based FPN} follows a design similar to the FPN in ViTDet \cite{DBLP:conf/eccv/LiMGH22}, which constructs multiscale feature maps exclusively from the final feature map of the vision backbone. Specifically, the DINOv2 (S/14) \cite{DBLP:journals/tmlr/OquabDMVSKFHMEA24} backbone produces a final feature map at a $\frac{1}{14}$ resolution, from which we generate feature maps at resolutions ${1, \frac{1}{4}, \frac{1}{14}, \frac{1}{28} }$. The highest resolutions ($1$ and $\frac{1}{4}$) are obtained via upsampling, followed by a single convolutional layer. The $\frac{1}{14}$ resolution is derived directly through a convolutional layer, while the $\frac{1}{28}$ resolution is obtained by applying a $2 \times 2$ max pooling operation prior to convolution. All convolutional layers employ $1 \times 1$ kernels, followed by normalization layers, primarily to adjust the channel dimensions of each feature map. The final multiscale feature maps have channel dimensions of $\{C, 2C, 4C, 8C\}$ for resolutions $\left\{1, \frac{1}{4}, \frac{1}{14}, \frac{1}{28}\right\}$, respectively, where $C = 128$.

As a baseline, we also adapted the FPN from SimpleClick \cite{DBLP:conf/iccv/0008XBN23}, which maps the feature map at $\frac{1}{14}$ resolution to scales $\left\{ \frac{2}{7}, \frac{1}{7}, \frac{1}{14}, \frac{1}{28}\right\}$. The resolutions $\frac{2}{7}$ and $\frac{1}{7}$ are generated using two and one transposed convolutional layers, respectively. The smallest resolution, $\frac{1}{28}$, is obtained by applying a convolutional layer with a $2 \times 2$ kernel and a stride of $2$. The feature map at the original resolution remains unchanged. At the final stage, all scales are processed by convolutional layers with a $1 \times 1$ kernel, followed by normalization layers. The output channel dimensions are consistent with those used in the Upsampler-based FPN.

For the multiscale segmentation head, we adopt the architecture from Segformer \cite{DBLP:conf/nips/XieWYAAL21}. Feature maps at four different scales are processed through independent convolutional layers with a $1 \times 1$ kernel and an output channel dimension of $256$. The features are then bilinearly interpolated to match the resolution of the largest feature map (either $1$ or $\frac{2}{7}$ in our setup), concatenated along the channel dimension, and passed through an additional convolutional layer with the same output channels and a $1 \times 1$ kernel. Finally, a classification layer is applied. Normalization layers are included as intermediate steps in the process.

We conducted experiments using DINOv2 (S/14) \cite{DBLP:journals/tmlr/OquabDMVSKFHMEA24} with a symmetric patch embedding click encoder and early injection. Models were trained on the SBD dataset \cite{DBLP:conf/iccv/HariharanABMM11} for 20 epochs with the batch size of 8. Similar to other experiments, VFM and upsampler modules are kept frozen. The results, shown in Tab. \ref{tab:multiscale_benchmark}, indicate that the Upsampler-based FPN, when combined with LoftUp \cite{huang2025loftuplearningcoordinatebasedfeature}, outperforms the one with bilinear interpolation. However, its advantage over the original FPN baseline remains inconclusive. Furthermore, our primary single-scale architecture consistently outperforms the multiscale approach. We hypothesize that the performance limitation arises from the fixed, predefined channel dimensions of feature maps, which are employed to maintain computational efficiency, leading to information loss in the upsampled features. One potential solution would be to retain all channels from the upsampled features, which could be explored in future works.

\medskip
\noindent \textbf{Multi-granular Benchmark.} A common challenge in IS is the ambiguity in determining the desired object scale based on a given input click, as the click may correspond to objects of varying granularities. Recently, GraCo \cite{DBLP:conf/cvpr/Zhao0CQZJL0C24} addressed this issue by introducing a granularity scale as an additional input parameter. The granularity scale, a continuous value between 0 and 1, specifies the intended object scale. To incorporate this concept, the authors extended the pre-trained SimpleClick model \cite{DBLP:conf/iccv/0008XBN23} by injecting learnable granularity embeddings into the segmentation pipeline. To enhance granularity control learning, LoRA fine-tuning \cite{DBLP:conf/iclr/HuSWALWWC22} was applied.

To construct our multi-granular IS benchmark, we closely follow the methodology introduced in GraCo \cite{DBLP:conf/cvpr/Zhao0CQZJL0C24}. Our approach builds upon our primary single-scale pipeline, which comprises VFM, click encoder, upsampler, and segmentation head. The click encoder and segmentation head are initialized using the results from previous experiments and, along with VFM and upsampler, remain frozen during training. Both the fixed granularity embeddings and the LoRA parameters are introduced as learnable components. The granularity embeddings are incorporated into the network by adding them element-wise to both image and click features, following either an early or late injection strategy. Simultaneously, LoRA fine-tuning is applied to \textbf{Q} and \textbf{K} projection layers in each attention block of the VFM, enabling efficient adaptation to multi-granular setup.

Tab. \ref{tab:multigranular_benchmark} presents the results for DINOv2 (S/14) using a symmetric patch embedding click encoder with early injection. The models were trained using the extended, multi-granular version of SBD dataset \cite{DBLP:conf/iccv/HariharanABMM11} for 20 epochs with the batch size of 16. The dataset generation, training, and evaluation protocols strictly follow those established by GraCo.

\input{sec/tables/multigranular_benchmark}

Here, we observe that the multi-granular setup performs considerably worse than the primary single-scale pipeline. Our hypothesis for this decline in performance is that applying LoRA fine-tuning to the backbone for integrating the new embeddings significantly distorts the backbone features, which serve as inputs to the upsampler, ultimately degrading the segmentation quality.

\section{Additional Visualizations}
We provide further visualizations of features after upsampling in Fig. \ref{fig:viz_feats_extra}, as well as segmentation masks after the first and third clicks in Figs. \ref{fig:viz_masks_extra_1st_click} and \ref{fig:viz_masks_extra_3rd_click}, respectively. All visualizations are generated on GrabCut dataset \cite{DBLP:journals/tog/RotherKB04} using the DINOv2 (S/14) \cite{DBLP:journals/tmlr/OquabDMVSKFHMEA24} backbone and a symmetric patch embedding click encoder with early injection.

\input{sec/figs/viz_feats_extra}
\input{sec/figs/viz_masks_extra_1st_click}
\input{sec/figs/viz_masks_extra_3rd_click}

%% file: sec/tables/multiscale_benchmark.tex
\begin{table*}[hb]
    \centering
    \renewcommand{\arraystretch}{1.1} % Adjust row height
    \setlength{\tabcolsep}{2.5pt} % Reduce column spacing
    \small
    \begin{tabular}{l|cccc!{\color{gray!50}\vrule}cccc!{\color{gray!50}\vrule}cccc}
        \toprule
        \multirow{2}{*}{\centering FPN} & \multicolumn{4}{c!{\color{gray!50}\vrule}}{GrabCut} & \multicolumn{4}{c!{\color{gray!50}\vrule}}{Berkeley} & \multicolumn{4}{c}{DAVIS} \\
         & NoC80 & NoC85 & NoC90 & IoU@1 & NoC80 & NoC85 & NoC90 & IoU@1 & NoC80 & NoC85 & NoC90 & IoU@1 \\
        \midrule
        SimpleClick FPN & \textbf{3.88}  & 5.52  & \textbf{9.74}  & \textbf{67.00}  & \textbf{5.18}  & \textbf{8.58}  & \textbf{13.61}  & \textbf{60.83}  & 10.81  & 14.25  & 17.51  & 54.50  \\
        Upsampler-based (Bilinear) & 4.84  & 7.02  & 11.16  & 61.42  & 6.29  & 10.11  & 15.81  & 55.60  & 11.80  & 15.08  & 17.88  & 53.63  \\
        Upsampler-based (LoftUp$^*$)  & 4.14  & \textbf{5.44}  & 10.16 & 63.01  & 5.26  & 8.85   & 13.64  & 59.39  & \textbf{10.40}  & \textbf{13.86}  & \textbf{17.20}  & \textbf{60.49}  \\
        \bottomrule
    \end{tabular}
    
    \caption{\textbf{Evaluation of the multiscale IS benchmark.} The backbone used is DINOv2 (S/14) \cite{DBLP:journals/tmlr/OquabDMVSKFHMEA24}, with a symmetric patch embedding click encoder and early injection. The segmentation head is an adapted version of SegFormer’s head \cite{DBLP:conf/nips/XieWYAAL21}. * Indicates results obtained from non-final checkpoints.}
    \label{tab:multiscale_benchmark}
\end{table*}

%% file: sec/tables/ablation_head.tex
\begin{table}[h]
    \centering
    \renewcommand{\arraystretch}{1.1} % Adjust row height
    \setlength{\tabcolsep}{2.5pt} % Reduce column spacing
    \small
    \begin{tabular}{l|cccc}
        \toprule
        \multirow{2}{*}{\centering Head} & \multicolumn{4}{c}{GrabCut} \\
         & NoC80 & NoC85 & NoC90 & IoU@1 \\
        \midrule
        Linear & 6.06 & 9.02 & 14.06 & 36.33 \\
        Simple Conv & 4.76 & 7.94 & 13.20 & 48.56 \\
        Conv & 4.40 & 6.14 & 9.80 & 56.03 \\
        \bottomrule
    \end{tabular}

    \caption{\textbf{Comparison of segmentation heads.} Models were trained on the SBD dataset for 20 epochs with a batch size of 16, using ViT \cite{DBLP:conf/iclr/DosovitskiyB0WZ21} backbone, a symmetric patch embedding click encoder with early injection, and no upsampler.}
    \label{tab:ablation_head}
\end{table}

%% file: sec/tables/multigranular_benchmark.tex
\begin{table}[h]
    \centering
    \renewcommand{\arraystretch}{1.1} % Adjust row height
    \setlength{\tabcolsep}{2.5pt} % Reduce column spacing
    \small
    \begin{tabular}{l|cccc}
        \toprule
        \multirow{2}{*}{\centering Upsampler} & \multicolumn{4}{c}{GrabCut} \\
         & NoC80 & NoC85 & NoC90 & IoU@1 \\
        \midrule
        Low-res & 5.28 & 6.92 & 10.80 & 59.39 \\
        FeatUp \cite{DBLP:conf/iclr/FuHBFZF24}  & \textbf{3.68} & \textbf{5.02} & \textbf{8.02} & \textbf{69.61}\\
        Low-res + GRA & 12.20 & 15.38 & 18.88 & 27.64  \\
        FeatUp \cite{DBLP:conf/iclr/FuHBFZF24} + GRA  & 9.78 & 11.28 & 13.92 & 12.30 \\
        \bottomrule
    \end{tabular}

    \caption{\textbf{Evaluation of the multi-granular IS benchmark.} The backbone used is DINOv2 (S/14) \cite{DBLP:journals/tmlr/OquabDMVSKFHMEA24}, with a symmetric patch embedding click encoder and early injection.}
    \label{tab:multigranular_benchmark}
\end{table}

%% file: sec/figs/viz_feats_extra.tex
% Use figure* for multi-column figure
\begin{figure*}[tp]
    \centering
    \includegraphics[width=0.95\linewidth]{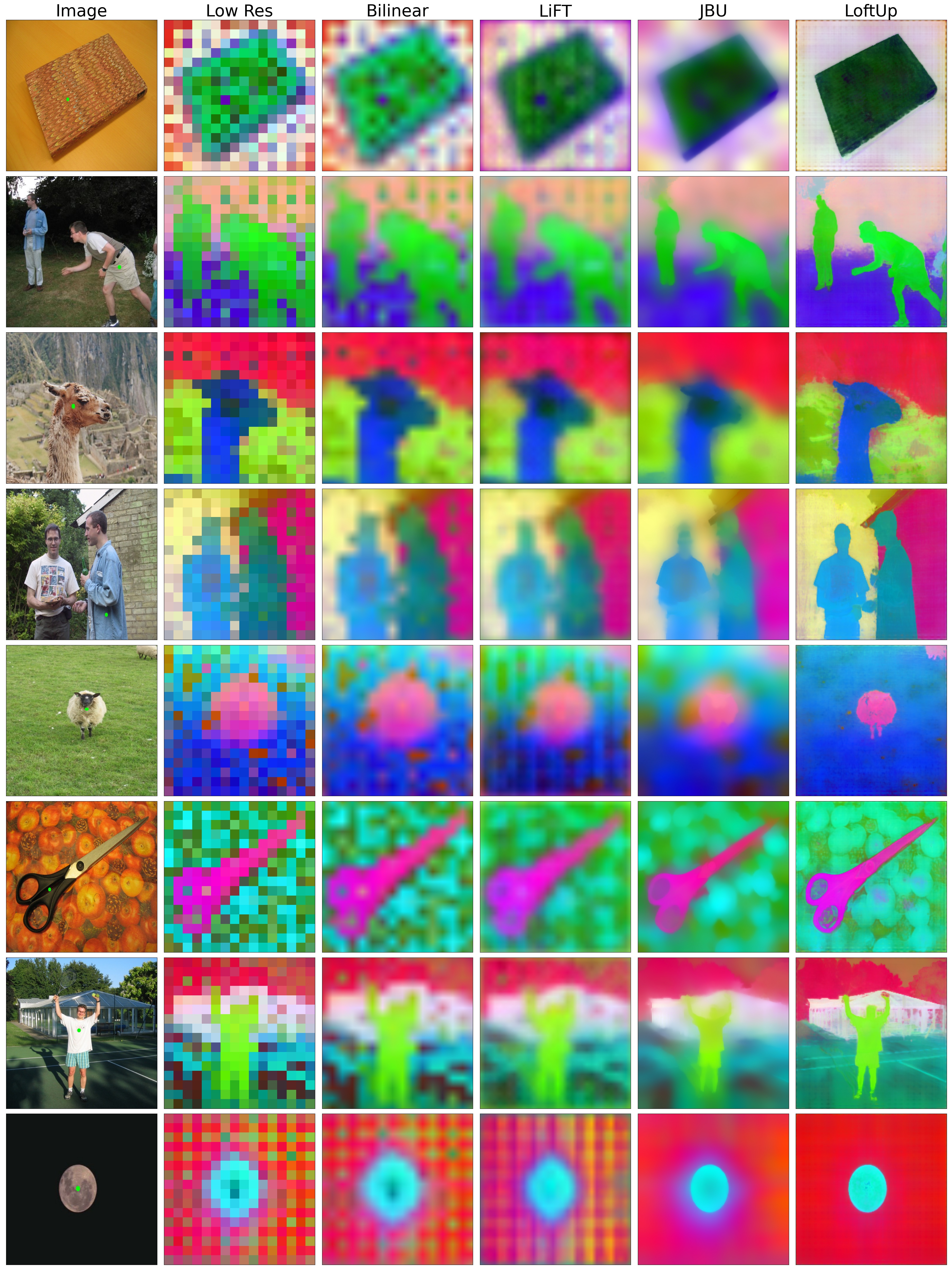}
\caption{\textbf{Additional visualizations of upsampler features with a single input click.} The click is indicated by a green dot on the original images.}
    \label{fig:viz_feats_extra}
\end{figure*}

%% file: sec/figs/viz_masks_extra_1st_click.tex
% Use figure* for multi-column figure
\begin{figure*}[tp]
    \centering
    \includegraphics[width=\linewidth]{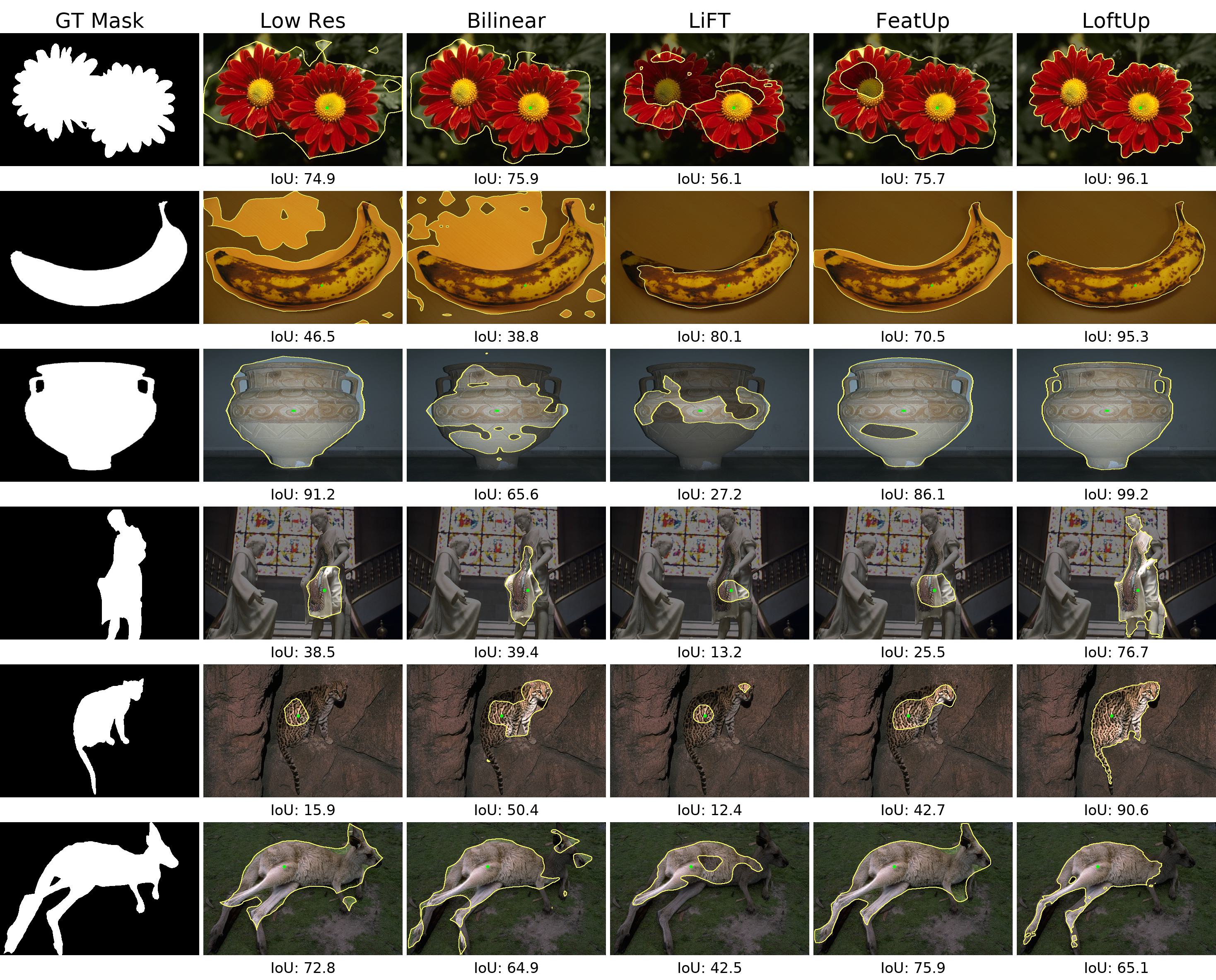}
    \caption{\textbf{Additional segmentation results with a single input click.} The click is indicated by a green dot.}
    \label{fig:viz_masks_extra_1st_click}
\end{figure*}

%% file: sec/figs/viz_masks_extra_3rd_click.tex
% Use figure* for multi-column figure
\begin{figure*}[tp]
    \centering
    \includegraphics[width=\linewidth]{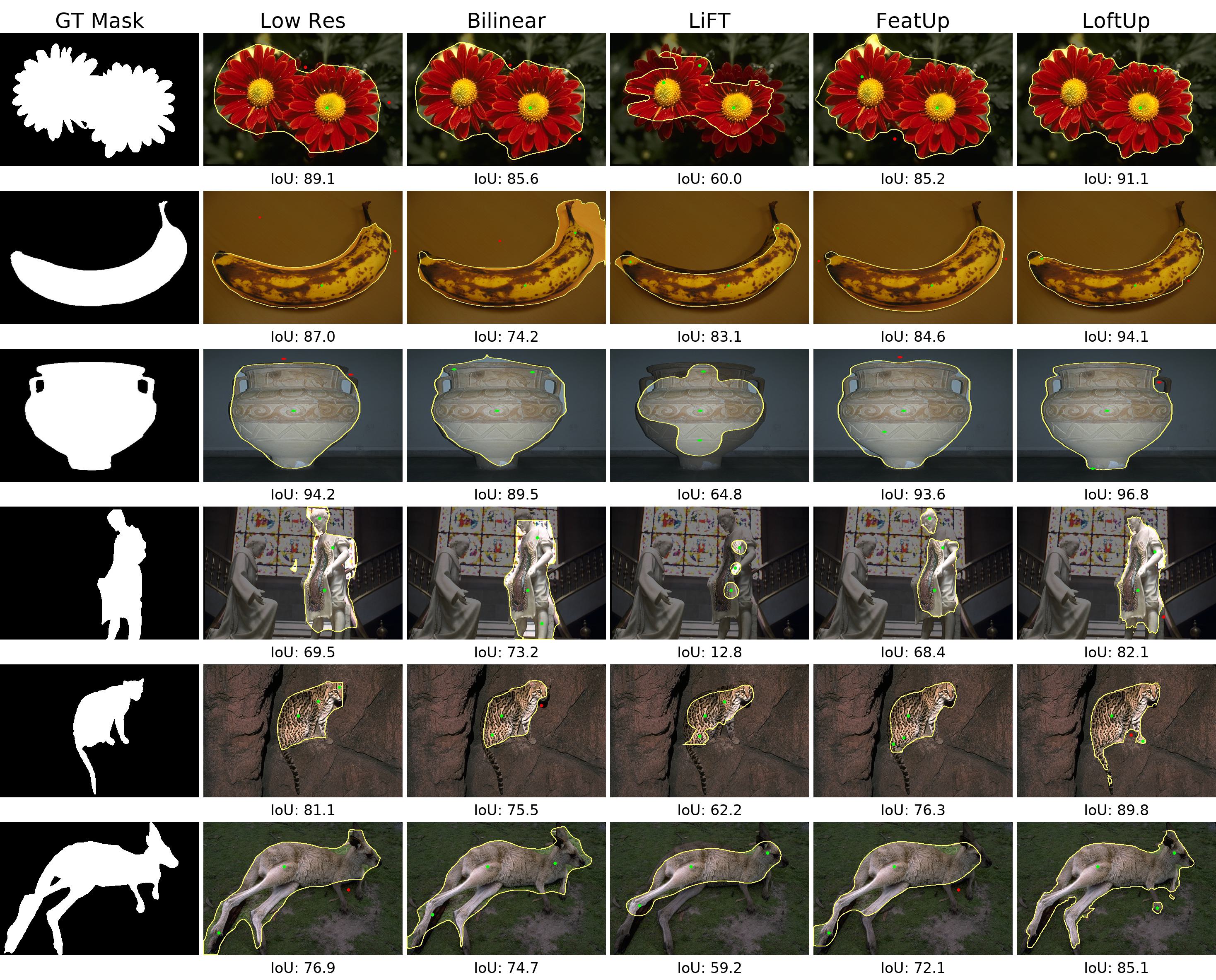}
    \caption{\textbf{Additional segmentation results with three input clicks.} Positive and negative clicks are indicated by green and red dots, respectively.}
    \label{fig:viz_masks_extra_3rd_click}
\end{figure*}